\def\year{2019}\relax
\documentclass[letterpaper]{article} 
\usepackage{aaai19}  
\usepackage{times}  
\usepackage{helvet}  
\usepackage{courier}  
\usepackage{url}  
\usepackage{graphicx}  
\usepackage{booktabs}
\usepackage{multirow}
\usepackage{subfigure}
\usepackage{amsmath}

\newcommand{\secref}[1]{Section \ref{#1}}
\newcommand{\figref}[1]{Figure \ref{#1}}

\newcommand{\tabref}[1]{Table \ref{#1}}
\newcommand{\bi}[1]{\textbf{\textit{#1}}}

\title{Automatic Generation of Chinese Short Product Titles for Mobile Display}

\author{
	Yu Gong,\textsuperscript{\rm 1}\footnotemark[1]
	Xusheng Luo,\textsuperscript{\rm 1}\footnotemark[1]
	Kenny Q. Zhu,\textsuperscript{\rm 2}
	Wenwu Ou,\textsuperscript{\rm 1}
	Zhao Li,\textsuperscript{\rm 1}
	Lu Duan\textsuperscript{\rm 3}
	\\ 
	\textsuperscript{\rm 1}Search Algorithm Team, Alibaba Group \\
	\textsuperscript{\rm 2}Shanghai Jiao Tong University \\
	\textsuperscript{\rm 3}Artificial Intelligence Department, Zhejiang Cainiao Supply Chain Management Co., Ltd.
	\\
	\{gongyu.gy, lxs140564\}@alibaba-inc.com, kzhu@cs.sjtu.edu.cn, \\
	santong.oww@taobao.com, lizhao.lz@alibaba-inc.com, duanlu.dl@cainiao.com
}

\begin{document}

\maketitle

\begin{abstract}
This paper studies the problem of automatically extracting a short 
title from a manually written longer description of E-commerce products
for display on mobile devices. 
It is a new extractive summarization problem on short text inputs,
for which we  propose a feature-enriched network model, combining three 
different categories of features in parallel. 
Experimental results show that our framework significantly outperforms 
several baselines by a substantial gain of 4.5\%.
Moreover, we produce an extractive summarization dataset for E-commerce 
short texts and will release it to the research community.
\end{abstract}

\renewcommand{\thefootnote}{\fnsymbol{footnote}}
\footnotetext[1]{Equal contribution.}
\renewcommand{\thefootnote}{\arabic{footnote}}

\section{Introduction}

Mobile Internet is fast becoming the primary venue for E-commerce.
People have got used to browsing through collections of products and making
transactions on the relatively small mobile phone screens. 
All major E-commerce giants such as Amazon, 
eBay and Taobao offer mobile apps that are poised to supersede the conventional websites. 
\begin{figure}[th]
	\centering
	\includegraphics[angle=0, width=0.7\columnwidth]{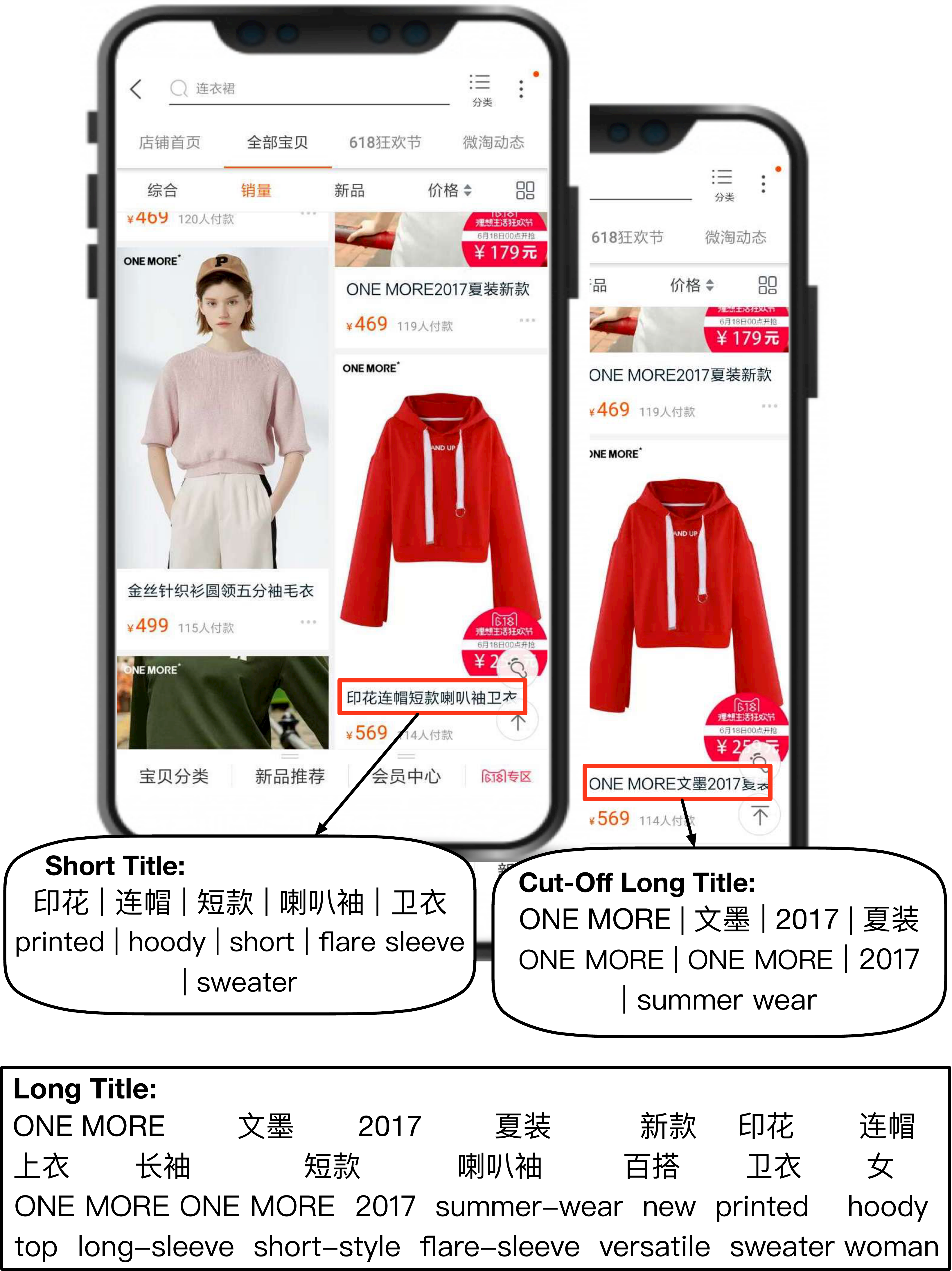}
	\caption{A cut-off long title on an E-commerce mobile app, vs. a corresponding
		short title.}
	\label{fig:compare}
\end{figure}

When a product is featured on an E-commerce website or mobile app, it is often
associated with a textual title which describes the key characteristics of
the product. These titles, written by merchants,
often contain gory details, so as to maximize
the chances of being retrieved by user search queries.
Therefore, such titles are often verbose, over-informative, and hardly 
readable. 
While this is okay for display on a computer's web browser, it becomes a
problem when such longish titles are displayed on mobile apps.
Take \figref{fig:compare} as an example. 
The title for a red sweater on an E-commerce mobile app is 
\includegraphics[height=1.3\fontcharht\font`\B]{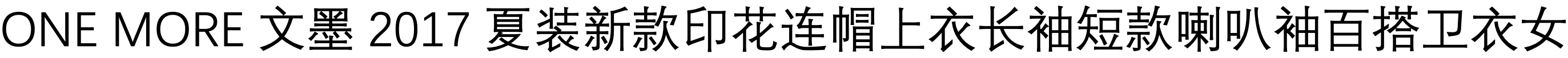}.
Due to the limited display space on mobile phones, 
original long titles (usually more than 20 characters) will be cut off, 
leaving only the first several characters 
``\includegraphics[height=1.3\fontcharht\font`\B]{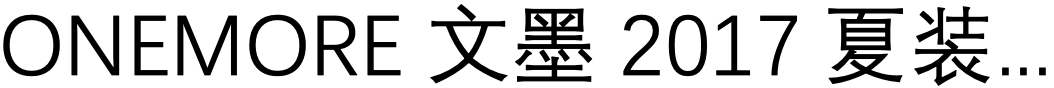}'' (ONE MORE 2017 summer woman...)
on the screen, which is completely incomprehensible, unless the user
clicks on the product and load the detailed product page.  

Thus, in order to properly display product listing on a mobile screen,
one has to significantly simplify (e.g., to under 10 characters) 
the long titles while keeping the most important information.
This way, user only has to glance through the search result page
to make quick decision whether they want to click into a particular 
product.
\figref{fig:compare} also shows as comparison an alternate display
of a shortened title for the same product.
The short title in the left snapshot is 
``\includegraphics[height=1.3\fontcharht\font`\B]{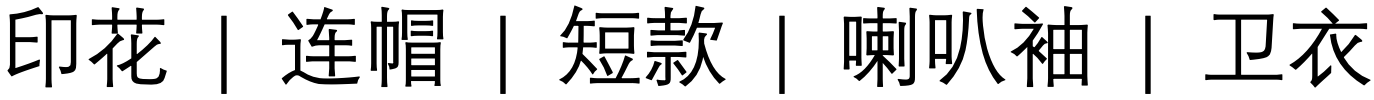}'',
which means ``printed$|$hoody$|$short$|$flare sleeve$|$sweater''. 

In this paper, we attempt to extract short titles from 
their longer, more verbose counterparts for E-commerce products.
To the best of our knowledge, this is the first attempt that attacks the 
E-commerce product short title extraction problem.

This problem is related to text summarization, 
which generates a summary by either {\em extracting} or {\em abstracting}
words or sentences from the input text. 
Existing summarization methods have primarily been applied to news or 
other long documents, which may contain irrelevant information. 
Thus, the goal of traditional summarization is to identify
the most essential information in the input and condense
it into something as fluent and readable as possible. 

We would attack our problem with an \textit{extractive} summarization 
approach, rather than an {\em abstractive} one for these reasons.
First, our input title is relatively shorter and contains less noise,
(average 27 characters; see \tabref{tab:data}).
Some words in the long title may not be important but they are all relevant 
to the product. Thus, it is sufficient to decide if each word should or 
should not stay in the summary. 
Second, the number of words in the output is strictly constrained in 
our problem due to size of the display. 
Generative (abstractive) approaches do not perform as well when there's
such a constraint.
Finally, for E-commerce, it is better for the words in the summary to come 
from the original title. 
Using different words may lead to a change of original intention of
the merchant.

State-of-the-art neural summarization models
\cite{cheng2016neural,narayan2017neural}
are generally based on attentional RNN frameworks and  have been applied on news or wiki-like articles.
However, in E-commerce, 
customers are not so sensitive to the order of the words in a product title. 
Besides using deep RNN with attention mechanism to encode word sequence, 
we believe other single-word level semantic features such as NER tags and 
TF-IDF scores will be as just as useful and should be given more weights in the model.
In this paper, we propose a feature-enriched neural network model, 
which is not only deep but also wide, 
aiming to effectively shorten original long titles.

The contributions of this paper are summarized below:
\begin{itemize}
	\item We collect and will open source a product title summary 
	dataset (\secref{sec:data}).
	\item We present a novel feature-enriched network model, 
	combining three different types of word level features
	(\secref{sec:model}), and the results 
	show the model outperforms 
	several strong baseline methods, with $ROUGE\textendash1_{F1}$ score of 
	0.725 (\secref{sec:eval_offline}). 
	\item By deploying the framework on an E-commerce mobile app, 
	we witnessed improved online sales 
	and better turnover conversion rate
	in the popular 11/11 shopping season (\secref{sec:eval_online}).
\end{itemize}

\section{Data Collection}
\label{sec:data}

Publicly available large-scale summarization dataset is rare.
Existing document summarization datasets include
DUC2\footnote{\url{http://duc.nist.gov/data.html}}, 
TAC3\footnote{\url{http://www.nist.gov/tac/2015/KBP/}} and 
TREC4\footnote{\url{http://trec.nist.gov/}} for English,
and LCSTS\footnote{\url{http://icrc.hitsz.edu.cn/Article/show/139.html}} for
Chinese.
In this work, we create a dataset on short title extraction for 
E-commerce products. 
This dataset comes from a module in \textbf{Taobao} named 
``Youhaohuo''
\footnote{\url{https://h5.m.taobao.com/lanlan/index.html}}.
{\em Youhaohuo} is a collection of high-quality 
products on Taobao.  If you click a product in Youhaohuo, you will 
be redirected to the detailed product page (including product title).
What is different from ordinary Taobao products is that online merchants are 
required to submit a short title for each Youhaohuo product.
This short title, written by humans, is readable and describes the 
key properties of the product.
Furthermore, most of these short titles are directly extracted from the 
original product titles. 
Thus, we believe Youhaohuo is a good data source of 
extractive summarization for product descriptions.
\begin{figure}[h]
	\centering
	\includegraphics[angle=0, width=1\columnwidth]{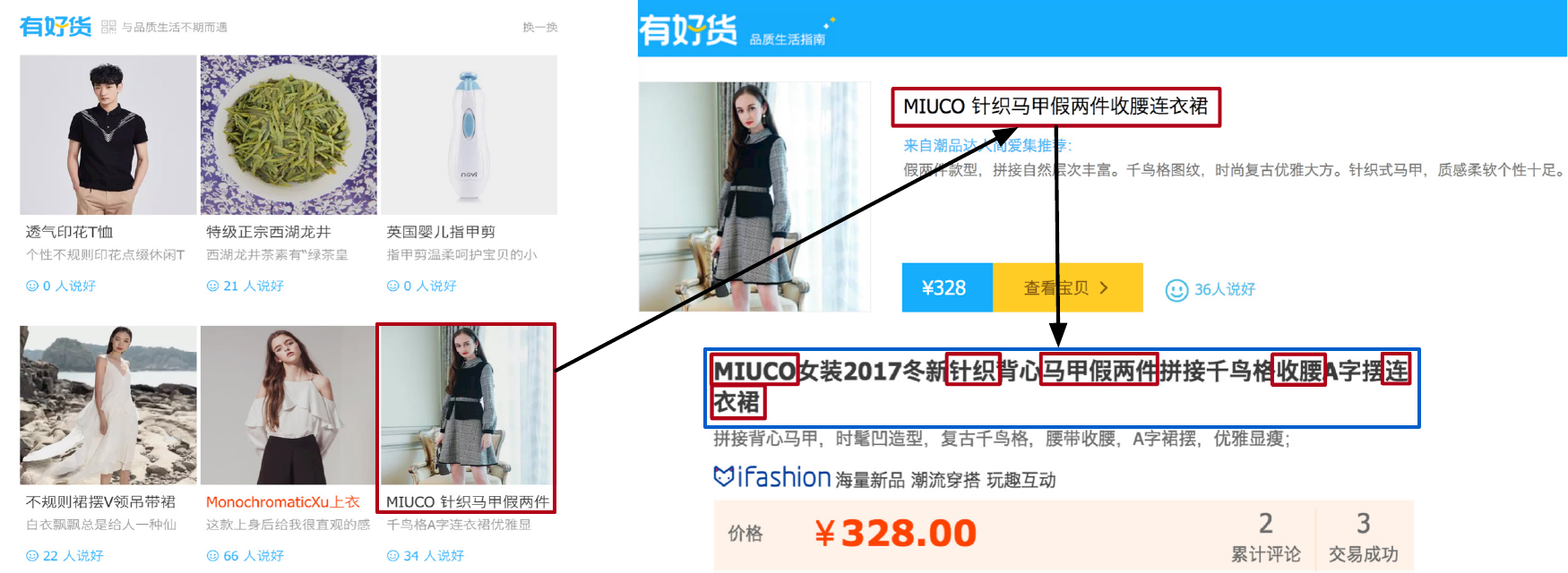}
	\caption{Procedure of data collection in Youhaohuo.}
	\label{fig:youhaohuo_demo}
\end{figure}

\figref{fig:youhaohuo_demo} shows how we collected the data.
On the left is a web page in Youhaohuo displaying several products, 
each of which contains an image and a short title below. 
When clicking on the bottom right dress, we jump to the detailed page 
on the right.  The title next to the picture in red box is the manually 
written short title, which says 
``\includegraphics[height=1.3\fontcharht\font`\B]{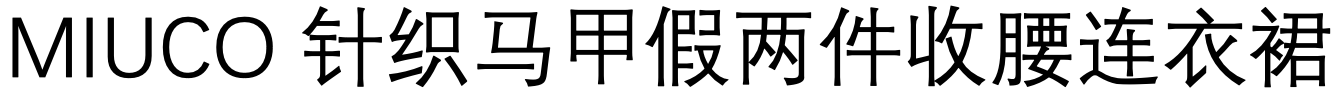}''
(MIUCO tight dress with knit vest).
This short title is extracted from the long title below in the blue box. 
Notice that all the characters in the short tile are directly extracted from 
the long title (red boxes inside blue box). 
In addition to the characters in the short title, the long title also 
contains extra information such as 
``\includegraphics[height=1.3\fontcharht\font`\B]{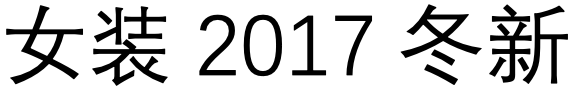}''
(woman's wear brand new in winter 2017). In this work, we segment the original long titles and 
short titles into Chinese words by jieba\footnote{\url{https://pypi.python/pypi/jieba/}}.

The dataset consists of 6,481,623 pairs of original and short product titles, 
which is the largest short text summarization dataset to date.
We call it large extractive summary dataset for E-commerce (LESD4EC),
whose statistics is shown in \tabref{tab:data}.
We believe this dataset will contribute to the future research of
short text summarization\footnote{The data can be found at 
	\url{https://github.com/pangolulu/product-short-title}. 
	Notice all the original data can be crawled online.}.
\begin{table}[th]
	\centering
	\scriptsize
	\label{tab:data}
	\begin{tabular}{l|r}
		\toprule
		No. of summaries & 6,481,623 \\
		\midrule
		No. of words per text  & 12 \\
		\midrule
		No. of chars per text & 27  \\
		\midrule
		No. of words per summary & 5  \\
		\midrule
		No. of chars per summary & 11  \\
		\bottomrule
	\end{tabular}
	\caption{Statistics of LESD4EC dataset.}
\end{table}

\section{Problem Definition}
\label{sec:problem}

In this section we formally define the problem of short title extraction.
A char is a single Chinese or English character.
A segmented word (or term) $x$ is a sequence of several chars such as 
``Nike'' or ``\includegraphics[height=1.3\fontcharht\font`\B]{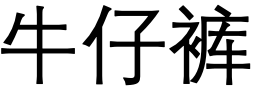}'' (jean).
A product title, denoted as $X$, is a sequence of words $\{x_1, x_2, ..., x_n\}$.
Let $Y$ be a sequence of labels $\{y_1, y_2, ..., y_n\}$ over $X$, where $y_i \in \{0, 1\}$.
The corresponding short title is a subsequence of $X$, denoted as $S = \{x_i\}$, 
where $y_i = 1$ and $|S| \le n$.

We regard short title extraction task as a sequence classification problem.
Each word is sequentially visited in the original product title order
and a binary decision is made.
We do this by scoring each word $x_i$ within $X$ and predicting a label $y_i \in \{0, 1\}$, 
indicating whether the word should or should not be included in the short title $S$.
As we apply supervised training, the objective is to maximize the likelihood of all word labels
$Y=\{y_1,y_2,...,y_n\}$, given the input product title $X$ and model parameters $\theta$:
\begin{equation}
\label{eqn:problem}
\log{p(Y|X,\theta)}=\sum_{i=1}^{n}{\log{p(y_i|X,\theta)}}.
\end{equation}

\section{Feature-Enriched Neural Extractive Model}
\label{sec:model}
\begin{figure}[h]
	\centering
	\includegraphics[angle=0, width=0.8\columnwidth]{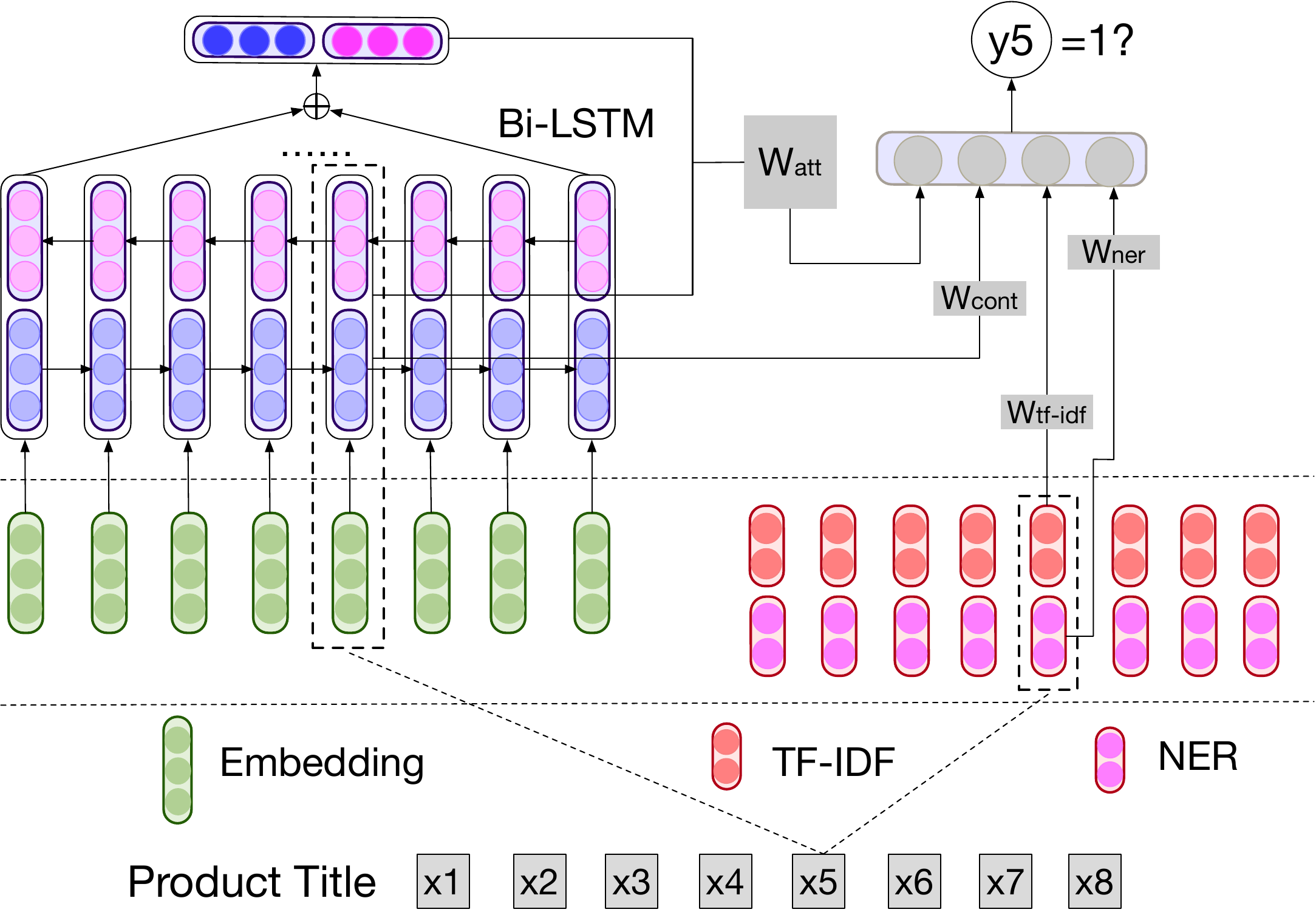}
	\caption{Architecture of Feature-Enriched Neural Extractive Model.}
	\label{fig:overview}
\end{figure}

In this section, we describe our extractive model for product short title extraction. 
The overall architecture of our neural network based extractive model is shown in \figref{fig:overview}.
Basically, we use a Recurrent Neural Network (RNN) as the main building block of the sequential classifier. 
However, unlike traditional RNN-based sequence labeling models used in NER or 
POS tagging, where all the word level features are fed into RNN cell, 
we instead divide the features into three parts, 
namely \emph{Content}, \emph{Attention} and \emph{Semantic} respectively. 
Finally we combine all three features in an ensemble.

\subsection{Content Feature}

To encode the product title, we first look up an embedding matrix $E_x\in \textbf{R}^{d\times v}$ to get the word embeddings $\bi{X}=(\bi{x}_1, \bi{x}_2, \dots, \bi{x}_n)$. 
Here, $d$ denotes the dimension of the embeddings and $v$ denotes the vocabulary size of natural language words. Then, the embeddings are fed into a bidirectional LSTM networks. 
To this end, we get two hidden state sequences, $(\overrightarrow{\bi{h}_1}, \overrightarrow{\bi{h}_2}, \dots, \overrightarrow{\bi{h}_n})$ from the forward network and $(\overleftarrow{\bi{h}_1}, \overleftarrow{\bi{h}_2}, \dots, \overleftarrow{\bi{h}_n})$ from the
backward network. 
We concatenate the forward hidden state of each word with corresponding backward hidden state, 
resulting in a representation $[\overrightarrow{\bi{h}_i};\overleftarrow{\bi{h}_i}]$. 
At this point, we obtain the representation of the product title $X$.

The \textit{content} feature of current word $x_i$ is then calculated as:
\begin{equation}
\label{eqn:content}
\bi{f}_{cont} = \bi{W}_{cont}^T [\overrightarrow{\bi{h}_i};\overleftarrow{\bi{h}_i}] + \bi{b}_{cont},
\end{equation}
where $\bi{W}_{cont}$ and $\bi{b}_{cont}$ are model parameters.

\subsection{Attention Feature}

In order to measure the importance of each word relevant to the whole product title, we borrow the idea of attention mechanism \cite{bahdanau2014neural} to calculate a relevance score between the hidden vector of current word and representation of the entire title sequence.

The representation of the entire product title is modeled as a non-linear transformation of the average pooling of the concatenated hidden states of the BiLSTM:
\begin{equation}
\bi{d}=\tanh{(\bi{W}_d{\frac{1}{n}}\sum_{j=1}^{n}{[\overrightarrow{\bi{h}_i};\overleftarrow{\bi{h}_i}]}+\bi{b}_d)}.
\end{equation}

Therefore, the attention feature of current word $x_i$ is calculated by a Bilinear combination function as:
\begin{equation}
\label{eqn:attention}
\bi{f}_{att} = \bi{d}^T\bi{W}_{att} [\overrightarrow{\bi{h}_i};\overleftarrow{\bi{h}_i}] + \bi{b}_{att},
\end{equation}
where $\bi{W}_{att}$ is a parameter matrix.

\subsection{Semantic Feature}

Apart from the two hidden features calculated using an RNN encoder, 
we design another kind of feature including TF-IDF and NER Tag, 
to capture the deep semantics of each word in a product title.

\subsubsection{TF-IDF}

Tf-idf, short for term frequency–inverse document frequency, 
is a numerical statistic that is intended to reflect how important a word is to a document in corpus or a sentence in a document.

A simple choice to calculate term frequency of current word $x_i$ is to use the 
number of its occurrences (or count) in the title:
\begin{equation}
\label{eqn:tf}
\textbf{tf}(x_i, X) =\frac{cnt_{x_i}}{n}.
\end{equation}

In the case of inverse document frequency, we calculate it as:
\begin{equation}
\label{eqn:idf}
\textbf{idf}(x_i) =\log(1+\frac{N}{C_{x_i}}),
\end{equation}
where $N$ is the number of product titles in the corpus and $C_{x_i}$ is the number of titles containing the word $x_i$.

By combining the above two, the tf-idf score of word $x_i$ in a product title $X$, denoted as $\textbf{tf-idf}(x_i, X)$, is then calculated as the product of $\textbf{tf}(x_i, X)$ and $\textbf{idf}(x_i)$.

We design a feature vector containing three values: tf score, idf score and tf-idf score and the calculate a third feature:
\begin{equation}
\label{eqn:tf-idf}
\bi{v}_{tfidf} = [\textbf{tf}(x_i, X); \textbf{idf}(x_i); \textbf{tf-idf}(x_i, X)]
\end{equation}
\begin{equation}
\bi{f}_{tfidf} =\bi{W}_{tfidf}^T \bi{v}_{tfidf} + \bi{b}_{tfidf}.
\end{equation}

\subsubsection{NER}   
We use a specialized NER tool for E-commerce to label entities in 
a product title. In total there are $36$ types of entities, which are of 
common interest in E-commerce scenario, such as ``color'', 
``style'' and ``size''.
For example, in segmented product title 
``\includegraphics[height=1.3\fontcharht\font`\B]{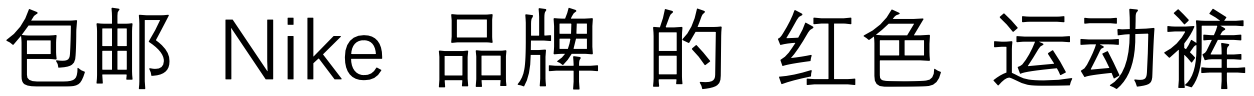}''
(Nike red sweatpants with free shipping),
``free shipping'' is labeled as \emph{Marketing\_Service},
``Nike'' is labeled as \emph{Brand}, 
``red'' is labeled as \emph{Color}
and ``sweatpants'' is labeled as \emph{Category}.
We use one-hot representation to encode NER feature $\bi{v}_{ner}$ of each word and 
then integrate it into the model by proposing a fourth feature:
\begin{equation}
\label{eqn:ner}
\bi{f}_{ner} = \bi{W}_{ner}^T \bi{v}_{ner} + \bi{b}_{ner}.
\end{equation}

\subsection{Ensemble}
\label{sec:inference}
We combine all the features above into one final score of word $x_i$:
\begin{equation}
\label{eqn:ner}
score_i = \sigma(\bi{W}_{f}^T [\bi{f}_{cont};\bi{f}_{att};\bi{f}_{tfidf};\bi{f}_{ner}] + \bi{b}_{f}),
\end{equation}
where $\sigma$ is the sigmoid or logistic function, which restrain the score between $0$ and $1$.
Based on that, we set a threshold $\tau$ to decide whether we keep word $x_i$ in the short title or not.

Our model is very much like the Wide \& Deep Model architecture \cite{cheng2016wide}.
While the \textit{content} and \textit{attention} features are deep since they rely on deep RNN structure,
the \textit{semantic} features are relatively wide and linear.

\section{Experiments}
In this section, we first introduce the experimental setup and the
previous state-of-the-art systems as comparison to 
our own model known as Feature-Enriched-Net.
We then show the implementation details and the evaluation results before
giving some discussions on the results.

\subsection{Training and Testing Data}

We randomly select 500,000 product titles as our training data, 
and another 50,000 for testing.
Each product title $X$ is annotated with a sequence of binary label $Y$,
i.e., each word $x_i$ is labeled with $1$ (included in short title) 
or $0$ (not included in short title). Readers may refer to \secref{sec:data} 
for the details about how we collect the product titles
and their corresponding short titles.

\subsection{Baseline Systems}
Since there are no previous work that directly solves the short title 
extraction for E-commerce product, we select our baselines from 
three categories. The first one is traditional methods. 
We choose a keyword extraction framework known as 
TextRank \cite{mihalcea2004textrank}.  It first infers an importance 
score for each word within the long title by an algorithm similar to PageRank, 
then decides whether each word should or should not be kept in the short title 
according to the scores.

The second category is standard sequence labeling systems. 
We choose the system mentioned by 
Huang et al.\shortcite{huang2015bidirectional},
in which a multi-layer BiLSTM is used.
Compared to our system, it does not exploit the attention mechanism and 
any side feature information.
We substitute the Conditional Random Field (CRF) layer with 
Logistic Regression to make it compatible with our binary labeling problem.
We call this system BiLSTM-Net.

The last category of methods is attention-based frameworks, 
which use encoder-decoder architecture with attention mechanism.
We choose Pointer Network \cite{vinyals2015pointer} as a comparison and 
call it Pointer-Net.
During decoding, it looks at the whole sentence, calculates the attentional 
distribution and then makes decisions based on the attentional probabilities.

\subsection{Implementation Details}
We pre-train word embeddings used in our model 
on the whole product titles data plus an extra corpus called 
``E-commerce Product Recommended Reason'',
which is written by online merchants and is also extracted from YouHaoHuo
(\secref{sec:data}).
We use the Word2vec \cite{mikolov2013distributed} 
CBOW model with context window size $5$, negative sampling size $5$, 
iteration steps $5$ and hierarchical softmax $1$. 
The size of pre-trained word embeddings is set to $200$. 
For Out-Of-Vocabulary (OOV) words, embeddings are initialized as zero.
All embeddings are updated during training. 
For recurrent neural network component in our system, 
we used a two-layers LSTM network with unit size $512$.
All product titles are padded to a maximum sentence length of $15$.
We perform a mini-batch cross-entropy loss training with a batch size 
of $128$ sentences for $10$ training epochs.  We use Adam optimizer, 
and the learning rate is initialized with $0.01$.

\subsection{Offline Evaluation}
\label{sec:eval_offline}
To evaluate the quality of automatically extracted short titles,
we used ROUGE \cite{lin2003automatic} to 
compare model generated
short titles to manually-written short titles.
In this paper, we only report ROUGE-1, mainly because 
linguistic fluency and word order are not of concern in this task.
Unlike previous works in which ROUGE is a recall-oriented metric,
we jointly consider precision, recall and F1 score,
since recall only presents the ratio of 
the number of extracted words included in ground-truth short title
over the total number of words in ground-truth short title.
However, due to the limited display space on mobile phones, 
the number of words (or characters) of extracted short title itself 
should be constrained as well.
Thus, precision is also measured in our experiments.
And $ROUGE\textendash1_{F1}$ is considered a comprehensive 
evaluation metric:
\begin{small}
	\begin{eqnarray*}
		& ROUGE_{P}=\frac{|S\cap S_{human}|}{|S|},\\
		& ROUGE_{R}=\frac{|S\cap S_{human}|}{|S_{human}|},\\
		& ROUGE_{F1}=\frac{2\cdot ROUGE_{P}\cdot ROUGE_{R}}{ROUGE_{P}+ROUGE_{R}}.
	\end{eqnarray*}
\end{small}
Where $S_{human}$ is the manually written short title, 
and $|S\cap S_{human}|$ is the number of overlapping words appearing 
in short titles generated by model and humans.

Final results on the test set are shown in \tabref{tab:eval1}.
We can find that our method (Feature-Enriched-Net) outperforms the baselines 
on both precision and recall.
Our method achieves the best $0.725$ F1 score, which improves by a relative gain of 4.5\%.
Pointer-Net would achieve higher precision score 
for its attentive ability to the whole sentence and select the most important words.
Our method considers long-short memories, attention mechanism
and other abundant semantic features.
That is to say, our model has the ability to
extract the most essential words from original long titles
and make the short titles more accurate and comprehensive.

\begin{figure}[th]
	\centering
	\includegraphics[angle=0, width=1\columnwidth]{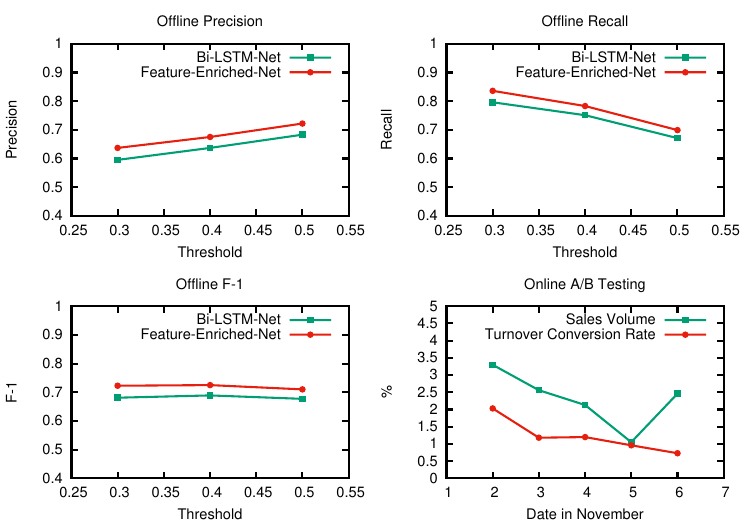}
	\caption{Offline results of Bi-LSTM-Net and Feature-Enriched-Net under different thresholds;
		Online A/B Testing of sales volume and 
		Turnover Conversion Rate}
	\label{fig:eval}
\end{figure}

We also tune the threshold $\tau$ used in BiLSTM-Net and our Feature-Enriched-Net.
This threshold indicates how large the predicted likelihood should be
so that a word will be included in the final short title.
We reported the results in \figref{fig:eval},
in which we set $\tau$ as $0.3$, $0.4$ and $0.5$.
From the figures, we can conclude that our model stably performs better than
the other.
\begin{table}[htbp]
	\centering
	\scriptsize
	\label{tab:eval1}
	\begin{tabular}{c|ccc}
		\toprule
		Models & $R\textendash1_{P}$ & $R\textendash1_{R}$ & $R\textendash1_{F1}$ \\
		\midrule
		TextRank & 0.430 & 0.219 & 0.290 \\
		BiLSTM-Net & 0.637 & 0.751 & 0.689 \\
		Pointer-Net & 0.648  & 0.746 & 0.694 \\
		Feature-Enriched-Net & \textbf{0.675} & \textbf{0.783} & \textbf{0.725} \\
		\bottomrule
	\end{tabular}
	\caption{
		Final results on the test set.
		We report ROUGE-1 Precision, Recall and corresponding F1.
		We use the tuned threshold $\tau=0.4$ (see \figref{fig:eval}) for BiLSTM-Net and Feature-Enriched-Net.
		Best ROUGE score in each column is highlighted in boldface.
	}
\end{table}

\subsection{Online A/B Testing}
\label{sec:eval_online}
This subsection presents the results of online evaluation in the search result page scenario
of an E-commerce mobile app with a standard A/B testing configuration.
Due to the limited display space on mobile phones,
only a fixed number of chars (12 characters in this app) 
can be shown out and excessive part will be cut off.
Therefore, unlike previously mentioned inference approach (with threshold $\tau$ in \secref{sec:inference}),
we regard it as classic Knapsack Problem.
Each word $x_i$ in the product title is an item with weight $len(x_i)$ and value $score_i$,
where $len(x_i)$ represents the char length of word $x_i$
and $score_i$ represents the predicted likelihood of word by our model.
The maximum weight capacity of our knapsack (also known as char length limit) is $m$. 
Then the target is:
\begin{equation*}
max \sum_{i=1}^{n} score_i z_i, \ \ s.t. \sum_{i=1}^{n} len(x_i)z_i \le m, z_i \in \{0, 1\}, 
\end{equation*}
where $z_i=1$ means $x_i$ should be reserved in the short title.
Similar to the standard solution to 0-1 Knapsack Problem,
we use a Dynamic Programming (DP) algorithm.

In our online A/B testing evaluation,
3\% of the users were randomly selected as testing group (about 3.4 million user views (UV)),
in which we substituted the original cut-off long title displayed to users
with extracted short titles by our model with DP inference.
We claim that after showing the short titles with most important keywords,
users have much better idea what the product is about on the 
``search result'' page and thus find the product they want more easily.
We deployed A/B testing for 5 days (from 2017-11-02 to 2017-11-06) 
and achieved on average \textbf{2.31\%} and \textbf{1.22\%} improvements of 
sales volume and turnover conversion rate (see \figref{fig:eval} for each day).
This clearly shows that better short product titles are more user-friendly
and hence improve the sales substantially.
\begin{table*}[ht]
	\scriptsize
	\centering
	\label{tab:case_study}
	\begin{tabular}{c|c}
		\toprule
		\multirow{2}{*}{ORIGINAL} & \includegraphics[height=2\fontcharht\font`\B]{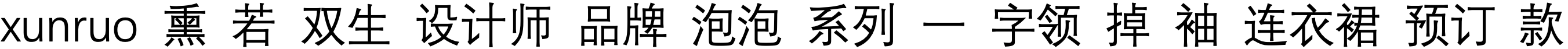}  \\
		& (Bookable XunRuo twin designer brand bubble series dress with boat neckline and off sleeves.) \\ 
		\midrule
		\multirow{2}{*}{HUMAN} & \includegraphics[height=2\fontcharht\font`\B]{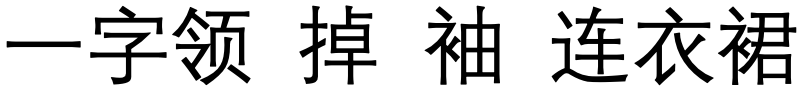} \\
		& (Dress with boat neckline and off sleeves.) \\
		\midrule
		\multirow{2}{*}{Feature-Enriched-Net} & \includegraphics[height=2\fontcharht\font`\B]{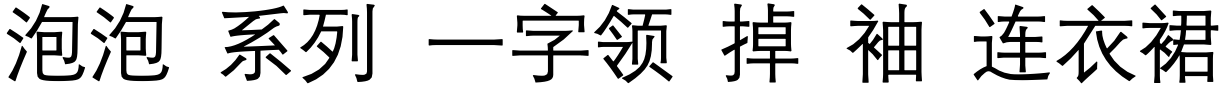} \\
		& (Bubble series dress with boat neckline and off sleeves.) \\
		\midrule
		\multirow{2}{*}{BiLSTM-Net} & \includegraphics[height=2\fontcharht\font`\B]{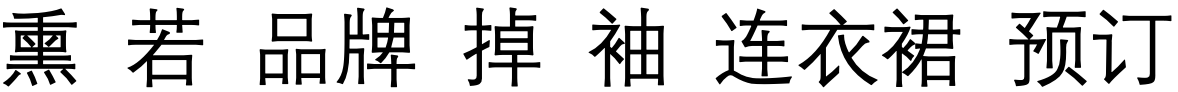}\\
		& (Bookable XunRuo brand dress with off sleeves.) \\
		\bottomrule
	\end{tabular}
	\caption{A real experimental case with 12-chars length limit.
		Pointer-Net is not included since encoder-decoder architecture can't directly adapt to character length limit}
\end{table*}

\subsection{Discussions}
In \tabref{tab:case_study}, we show a real case of original long title, 
along with short title annotated by human beings, 
predicted by BiLSTM-Net and Feature-Enriched-Net respectively.
From the human annotated short title,
we find that a proper short title should contain 
the most important elements of the product,
such as category (``dress'') and description of properties (``boat neckline and off sleeves'').
While some other elements such as 
brand terms (``XunRuo twin designer brand'')
or service terms (``open for reservation'') 
should not be kept in the short title.
Our Feature-Enriched Net has the ability to generate
a satisfying short title,
while baseline model tends to miss some essential information.

\begin{figure}[th]
	\centering
	\includegraphics[angle=0, width=0.8\columnwidth]{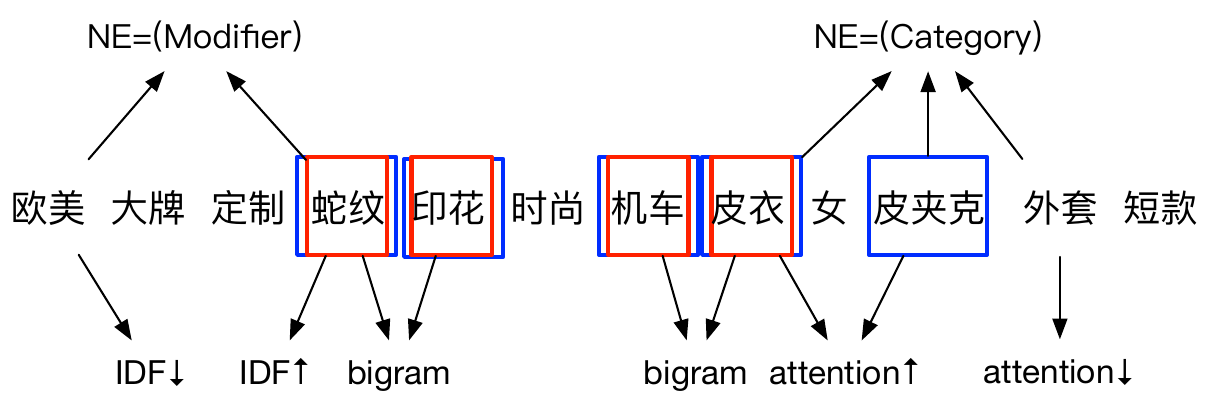}
	\caption{An example of product long title, ground-truth short title (with red boxes) and model predicted short title (with blue boxes). NE=(?) means the named entity type. $\uparrow$ means a higher score. $\downarrow
		$ means a lower score.}
	\label{fig:case_study}
\end{figure}

Furthermore, in order to visually see the performance of our Feature-Enriched-Net model,
we also carried out a feature analysis.
In the following example (see \figref{fig:case_study}): 
``\includegraphics[height=1.3\fontcharht\font`\B]{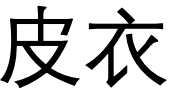}''(leather clothing), ``\includegraphics[height=1.3\fontcharht\font`\B]{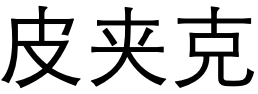}''(leather jacket), ``\includegraphics[height=1.3\fontcharht\font`\B]{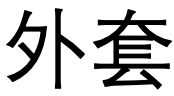}''(jacket) are all the category words (Named Entity type as \emph{Category}),
but the ``leather clothing'' and ``leather jacket'' have higher attention scores.
``\includegraphics[height=1.3\fontcharht\font`\B]{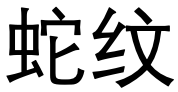}''(snake) and ``\includegraphics[height=1.3\fontcharht\font`\B]{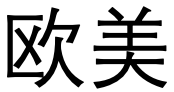}''(Europe and America) both belong to modifiers of the product,
but the IDF score of ``snake'' is much higher than ``Europe and America''.
By introducing BiLSTM,
the sequential model can learn that ``\includegraphics[height=1.3\fontcharht\font`\B]{shewen.png}''(snake) and ``\includegraphics[height=1.3\fontcharht\font`\B]{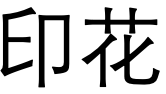}''(print) form a bigram pattern and they should appear together,
so as ``\includegraphics[height=1.3\fontcharht\font`\B]{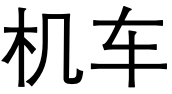}''(motorcycle) and ``\includegraphics[height=1.3\fontcharht\font`\B]{piyi.png}''(leather clothing).

However, there is still room for improvement.
Terms with similar meaning may co-occur
in the short title generated by our model,
when they all happen to be important terms such as a category.
For example, 
``\includegraphics[height=1.3\fontcharht\font`\B]{piyi.png}'' and ``\includegraphics[height=1.3\fontcharht\font`\B]{pijiake.png}'' both mean ``jacket'' in a long title,
and the model tends to keep both of them.
However, only one of them is enough and
the space saved can be used to display other useful information to customers.
We will explore intra-attention \cite{paulus2017deep} as an extra feature in our future work.

\section{Related Work}
Our work can be comfortably set in the area 
of \textit{short text summarization}. 
This line of research is in fact essentially sentence compression
working on short text inputs such as tweets, microblogs or single sentence. 
Recent advances typically contribute to improving seq-to-seq learning, 
or attentional RNN encoder-decoder structures \cite{nallapati2016abstractive}.
While these methods are mostly abstractive, 
we use an extractive framework,
combining with deep attentional RNN network and explicit semantic features, due to different scenarios of summarization problem.
The most related work is \cite{wang2018multi},
since they also try to compress product title in E-commerce. 
They use user search log as an external knowledge to guide the model and regard short title as some kind of query, 
to serve the purpose of improving online business values.
Differently, 
user search log is not necessary in our model,
and our goal is to make the short title as real as if it is written by human.

\section{Conclusion}
To the best of our knowledge, 
this is the first piece of work that focuses on 
extractive summarization for E-commerce product title.
We propose a deep and wide model, combining attentional RNN
framework with rich semantic features such as TF-IDF scores and NER tags.
Our model outperforms several popular summarization models,
and achieves ROUGE-1 F1 score of 0.725.
Besides, the result of online A/B testing shows substantial benefits
of our model in real online shopping scenario.
Possible future works include handling similar terms
that appear in the short titles generated by our model.

\bibliographystyle{aaai}
\bibliography{paper}

\end{document}